\documentclass[11pt,letterpaper]{article}
\usepackage[pass]{geometry}

\usepackage{fullpage}
\usepackage{authblk}
\usepackage[english]{babel}
\usepackage[utf8x]{inputenc}
\usepackage{amsmath}
\usepackage{amssymb}
\usepackage{graphicx}
\graphicspath{ {images/} }
\usepackage{longtable}
\usepackage{multirow}
\usepackage{hyperref}
\usepackage{subcaption}
\usepackage{comment}
\usepackage{mathtools}
\usepackage[table,xcdraw]{xcolor}
\title{RamanNet: A generalized neural network architecture for Raman spectrum analysis}
\author[1]{Nabil Ibtehaz}
\author[2,*]{Muhammad E. H. Chowdhury}
\author[2]{Amith Khandakar}
\author[2]{Serkan Kiranyaz}
\author[3]{M. Sohel Rahman}
\author[4,*]{Susu M. Zughaier}

\affil[1]{Department of Computer Science, Purdue University, West Lafayette, IN 47907, United States, nibtehaz@purdue.edu (NI)}
\affil[2]{Department of Electrical Engineering, Qatar University, Doha-2713, Qatar; mchowdhury@qu.edu.qa (MEHC), amitk@qu.edu.qa (AK), mkiranyaz@qu.edu.qa (SK)}
\affil[3]{Department of Computer Science and Engineering, Bangladesh University of Engineering and Technology, Dhaka-1205, Bangladesh,  msrahman@cse.ac.buet.bd (MSR)}
\affil[4]{Department of Basic Medical Sciences, College of Medicine, QU Health, Qatar University, Doha, Qatar, PO Box 2713, szughaier@qu.edu.qa (SMZ)}
\affil[*]{Corresponding author : Muhammad E. H. Chowdhury (mchowdhury@qu.edu.qa); Susu M. Zughaier (szughaier@qu.edu.qa)}

\date{}

\begin{document}

\maketitle

\begin{abstract}
Raman spectroscopy provides a vibrational profile of the molecules and thus can be used to uniquely identify different kind of materials. This sort of molecule fingerprinting has thus led to widespread application of Raman spectrum in various fields like medical diagnosis, forensics, mineralogy, bacteriology and virology etc. Despite the recent rise in Raman spectra data volume, there has not been any significant effort in developing generalized machine learning methods targeted towards Raman spectra analysis. We examine, experiment and evaluate existing methods and conjecture that neither current sequential models nor traditional machine learning models are satisfactorily sufficient to analyze Raman spectra. Both have their perks and pitfalls, therefore we attempt to mix the best of both worlds and propose a novel network architecture RamanNet. RamanNet is immune to the invariance property in convolutional neural networks (CNNs) and at the same time better than traditional machine learning models for the inclusion of sparse connectivity. Our experiments on 4 public datasets demonstrate superior performance over the much complex state-of-the-art methods and thus RamanNet has the potential to become the defacto standard in Raman spectra data analysis.
\end{abstract}

\textbf{Keywords}: Raman spectrum analysis, Convolutional Neural Networks, Multilayer perceptron, Deep learning, Neural network

\section{Introduction}

Raman scattering is one of the various light-matter interactions, comprising absorption and subsequent emission of light by matter \cite{gardiner1989introduction}. Spectroscopy, being the study of the interaction of light or broader electromagnetic radiation with matter, thus has also focused on Raman scattering, ever since it's discovery in 1928 by Raman and Krishnan \cite{raman1928new}. Unlike elastic scattering, e.g. Raleigh scattering, the wavelength of incident light changes in Raman scattering \cite{cornel2012raman}. Following the typical norm of inelastic light scattering, when a photon excites the sample, the electrons are raised to a higher virtual energy state \cite{hammes2005spectroscopy}. This excitation event is usually short-lived and the molecule soon reaches a new stable energy state, either lower (Stokes shift) or higher (anti-Stokes) \cite{jones2019raman}. Based on the difference in energy, the sample achieves a different vibrational and rotational state. Therefore, Raman spectroscopy can be used to analyze the vibrational modes of various molecules, extracting the structural fingerprint of such materials in the process \cite{gardiner1989introduction}.

Being capable of uniquely fingerprinting materials, Raman spectra have been used in a wide variety of applications \cite{lussier2020deep}, covering medical diagnosis \cite{wu2021highly}, forensics \cite{braz2013raman}, mineralogy \cite{liu2017deep}, bacteriology \cite{wu2014culture}, virology \cite{shanmukh2008identification}, etc.

Conventionally Raman spectra are analyzed in terms of wavenumber $\Tilde{v}$ $(cm^{-1})$. The standard practice is to present them with the wavenumber shift linearly increasing along the horizontal axis. On the contrary, the vertical axis ordinate is proportional to intensity \cite{jones2019raman}. Therefore, this setting does not appear too different from traditional spectograms. However, the issue in treating Raman spectra as typical spectrograms is that for Raman spectrum, we do not have time along the horizontal axis. Therefore, it is not logically sound to apply models that are used for spectrum analysis, e.g., convolutional neural networks (CNN), as they discard the (time-domain) locality of the spectrum, which is crucial for Raman spectra. On the contrary, although the traditional machine learning methods are more suitable to deal with Raman spectra, they suffer from the curse of dimensionality as Raman spectra data is usually long. Principal Component Analysis (PCA) has been widely used for feature reduction, as 34 out of recent 52 papers used that \cite{lussier2020deep}. Still it may not always be able to compress Raman spectra properly, as PCA is more suitable for tabular data source, where the features are ideally uncorrelated, but in Raman spectra the intensities at nearby Raman shifts are expected to demonstrate some sort of correlation.

With the reduction of cost and complexity related to the Raman data extraction pipeline, there has been an unprecedented expansion in Raman datasets in recent years. This sudden growth of available Raman data requires suitable methods to analyze them properly. However, to the best of our knowledge, there has not been any machine learning methodology, devised solely focusing on Raman spectra data, considering the pattern and properties of this unique data source. To this end, we carefully contemplate the properties of Raman spectra data and corresponding attributes of contemporary machine learning methods. We argue why the application of convolutional neural networks (CNN) may not be appropriate for the invariance properties, but at the same time acknowledge that CNN is more capable than traditional machine learning methods due to the sparse connection and weight sharing properties. We make an attempt to fuse the best of both worlds and propose RamanNet, a deep learning model that also follows the paradigm of sparse connectivity without the limitation of temporal or spatial invariance. We achieve this by using shifted densely connected layers and emulate sparse connectivity found in CNN, without the concern of invariance. Furthermore, dimensionality reduction has been involved in most Raman spectra applications, thus we employ triplet loss in our hidden layers and make more separable and informative embeddings in lower-dimensional space. Our experiments on 4 public datasets demonstrate superior performance over the much complex state-of-the-art methods and thus RamanNet has the potential to become the defacto standard in Raman spectra data analysis.

\section{Motivations and High-Level Considerations}

From a visual perception, the Raman spectrum resembles a signal-like waveform, which has led to the application of 1D Convolutional Neural Networks to analyze Raman spectra \cite{liu2017deep,ho2019rapid,erzina2020precise}. However, the Raman spectrum is not a typical signal rather it's an energy distribution plot, which may not be suitable for appropriate utilization of CNNs as our subsequent discussion unfolds. In what follows, we briefly discuss the properties of CNNs and present our motivations and rationale. 

Convolutional Neural Networks are one of the most successful and widely used neural network architectures, particularly in computer vision \cite{yoo2015deep} and signal processing \cite{kiranyaz20191} domains. The success of CNNs can be largely attributed to three primary properties of CNNs, namely sparse interaction, parameter sharing and equivariant representations \cite{Goodfellowetal2016}. The nature of image or signal data, i.e., spatial or temporal properties, works in perfect harmony with the properties of CNN. Unlike traditional feed-forward neural networks, where all the input features are connected with all the hidden or output layers, CNNs employ sparse connectivity by leveraging kernels. Only a small portion of the input is analyzed at a specific step using kernels, thus this prevents the computation from being overwhelmed with analyzing the entire input at once. Although the focus is put on local information, global information is also considered by appropriate use of pooling operations along with the broader field of vision in deeper networks. Furthermore, the use of kernel-based computation also faciliatates parameter sharing, reducing the number of parameters and risk of overfitting simultaneously. Another vital property of CNN is the equivariance in representation. Mathematically for a convolutional operation $f$ on an image $I$, $f(I(x,y)) = f((I(x-h,y-k))$ , i.e., a point of interest whether it resides at location $(x,y)$ or at a shifted location $(x-h,y-k)$, the output of the convoluation operation remains the same.

This equivarience to translation plays a pivotal role in working image or signal data. When processing signal data, this property implies that CNN generates a timeline of the emergence of different keypoints in the signal. As a result, regardless of the time of occurrence, all the features in a signal are captured, unlike traditional neural networks which would have only sought for the features at the exactly fixed timestamps. Furthermore, the application of global pooling operations makes sure that all the feature signatures are preserved in the final representation.

\begin{figure}[ht]
     \centering
     \begin{subfigure}[b]{0.325\textwidth}
         \centering
         \includegraphics[width=\textwidth]{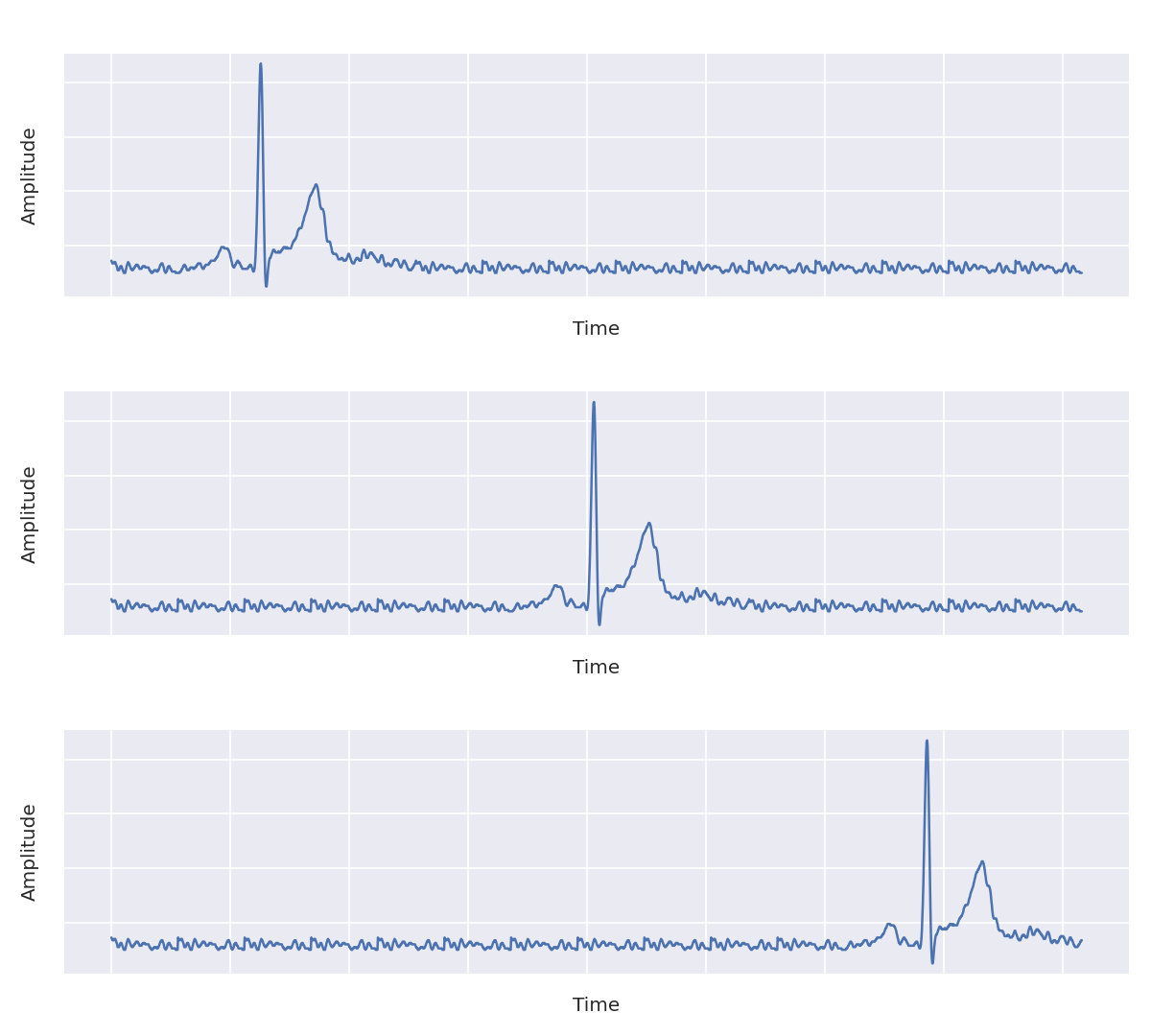}
         \caption{}
     \end{subfigure}
     \begin{subfigure}[b]{0.325\textwidth}
         \centering
         \includegraphics[width=\textwidth]{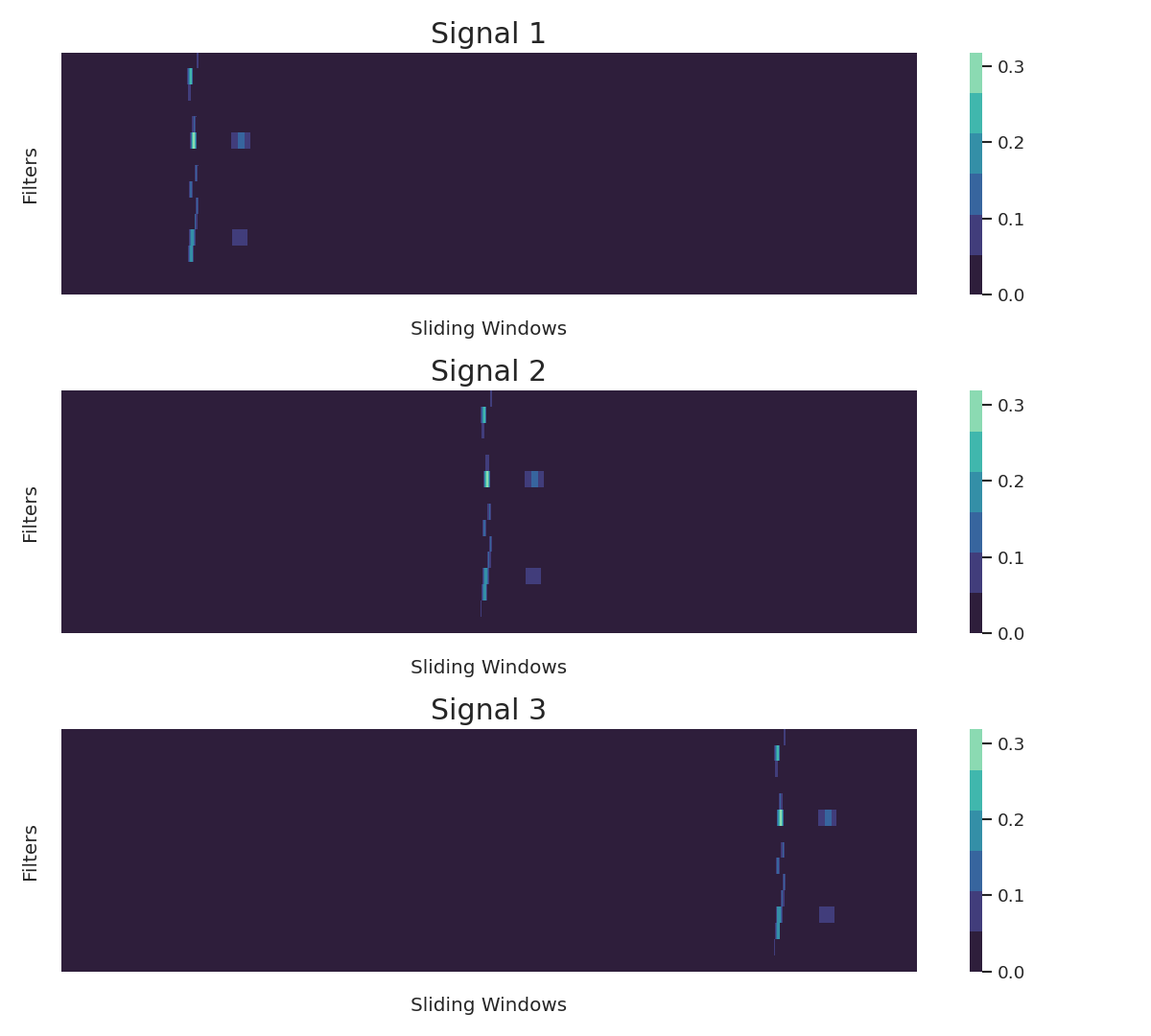}
         \caption{}
     \end{subfigure}
     \begin{subfigure}[b]{0.325\textwidth}
         \centering
         \includegraphics[width=\textwidth]{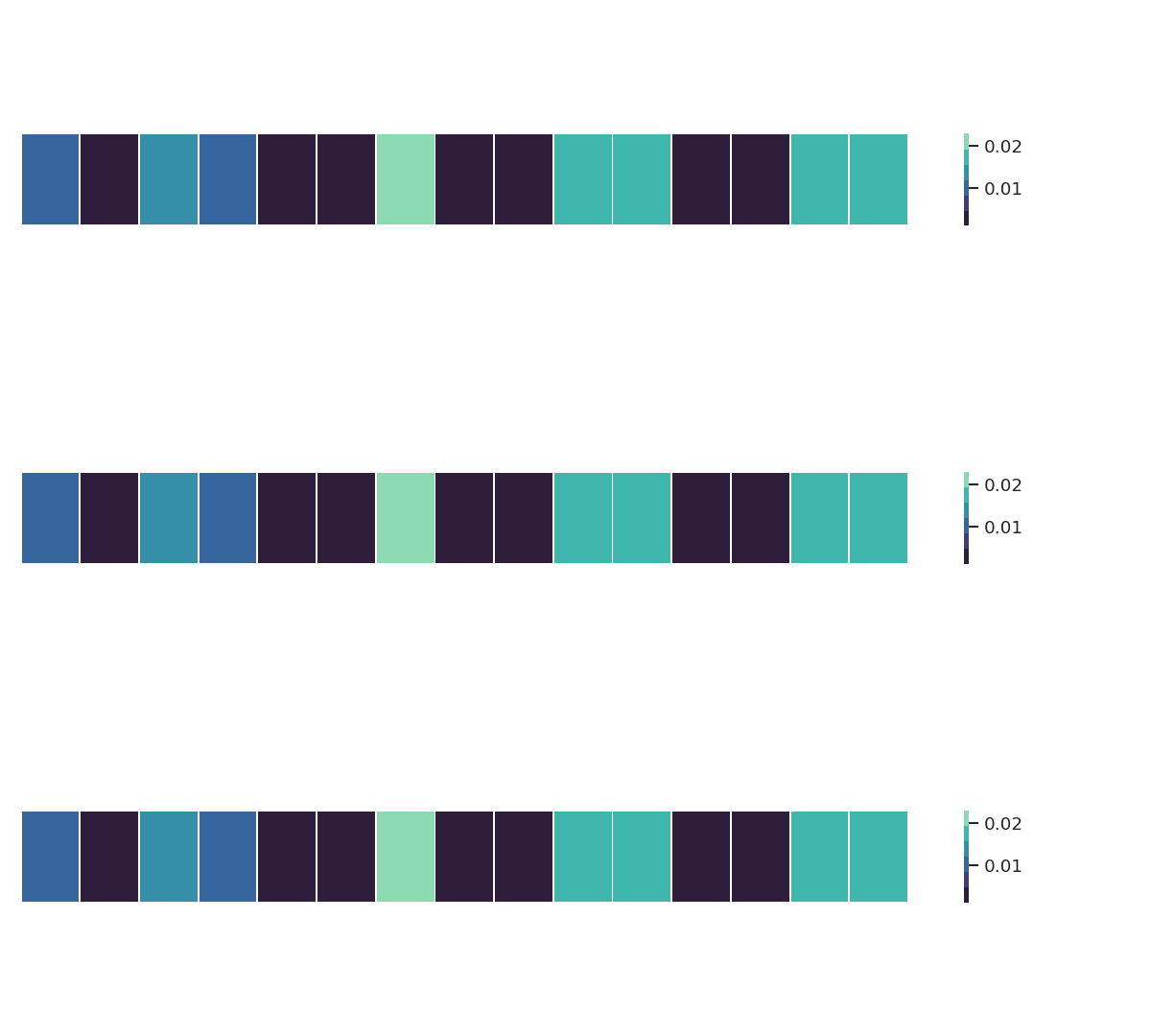}
         \caption{}
     \end{subfigure}
     \caption{Analysis of CNN for a 1-D time-series signal input.}
     \label{fig:cnn_sig}
\end{figure}

In Fig. \ref{fig:cnn_sig}, we have presented a simplified example of how CNN works with a signal as input. In Fig. \ref{fig:cnn_sig}a, we have three shifted ECG signals, since we cannot always be certain of the occurrence of a particular feature when dealing with signals, it is imperative that the model can identify the feature irrespective of the timestamp. In Fig. \ref{fig:cnn_sig}b, we can observe the featuremaps from a convoluational layer, it is apparent that the equivalent features are detected, albeit shifted in accordance with the shifted nature of the signals, complying with the property $f(I(x,y))=f(I(x-h,y-k))$. Finally, a global pooling operation summarizes the featuremaps and provides identical representations of all the three signals in Fig. \ref{fig:cnn_sig}c.

\begin{figure}[ht]
     \centering
     \begin{subfigure}[b]{0.325\textwidth}
         \centering
         \includegraphics[width=\textwidth]{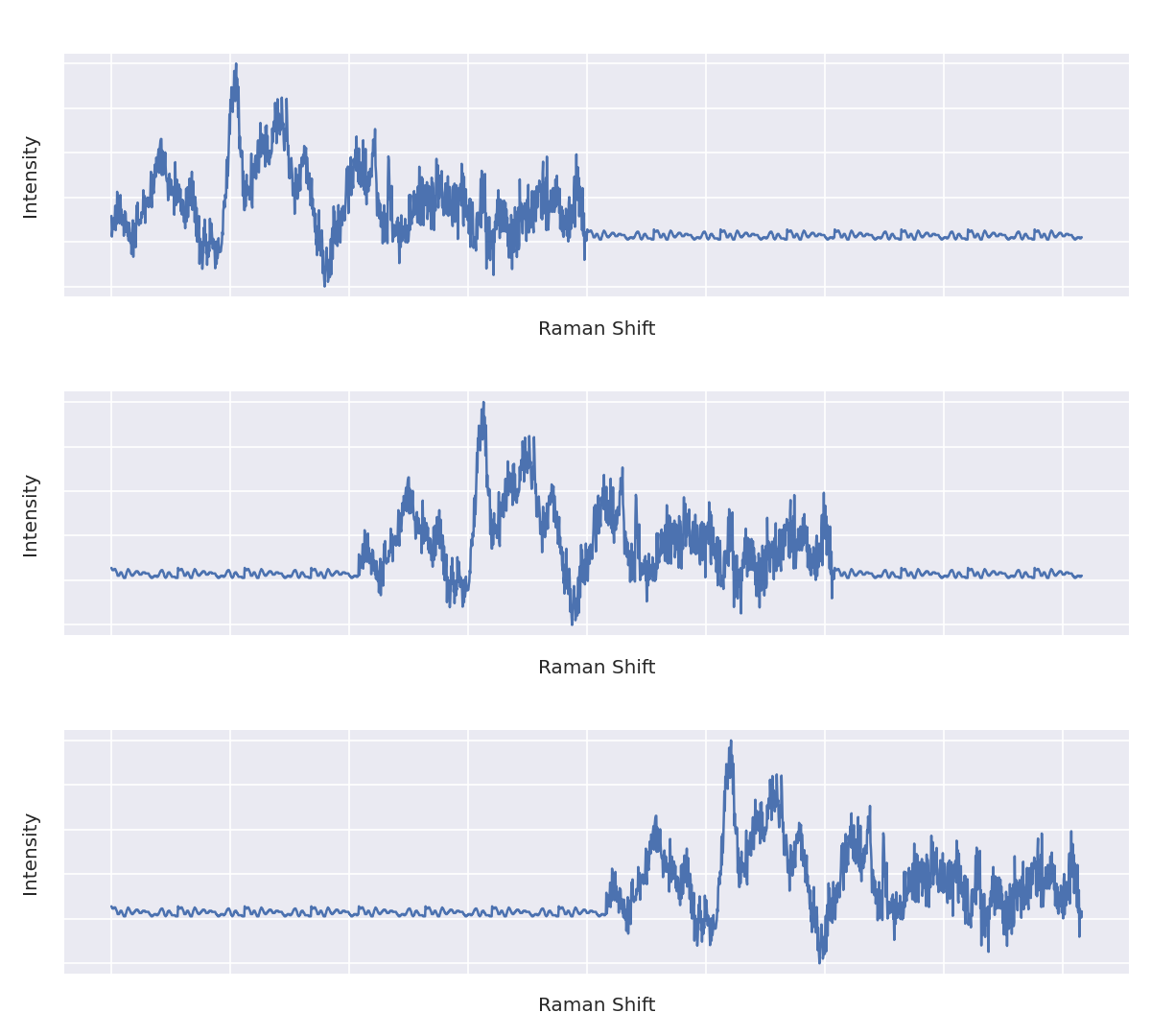}
         \caption{}
     \end{subfigure}
     \begin{subfigure}[b]{0.325\textwidth}
         \centering
         \includegraphics[width=\textwidth]{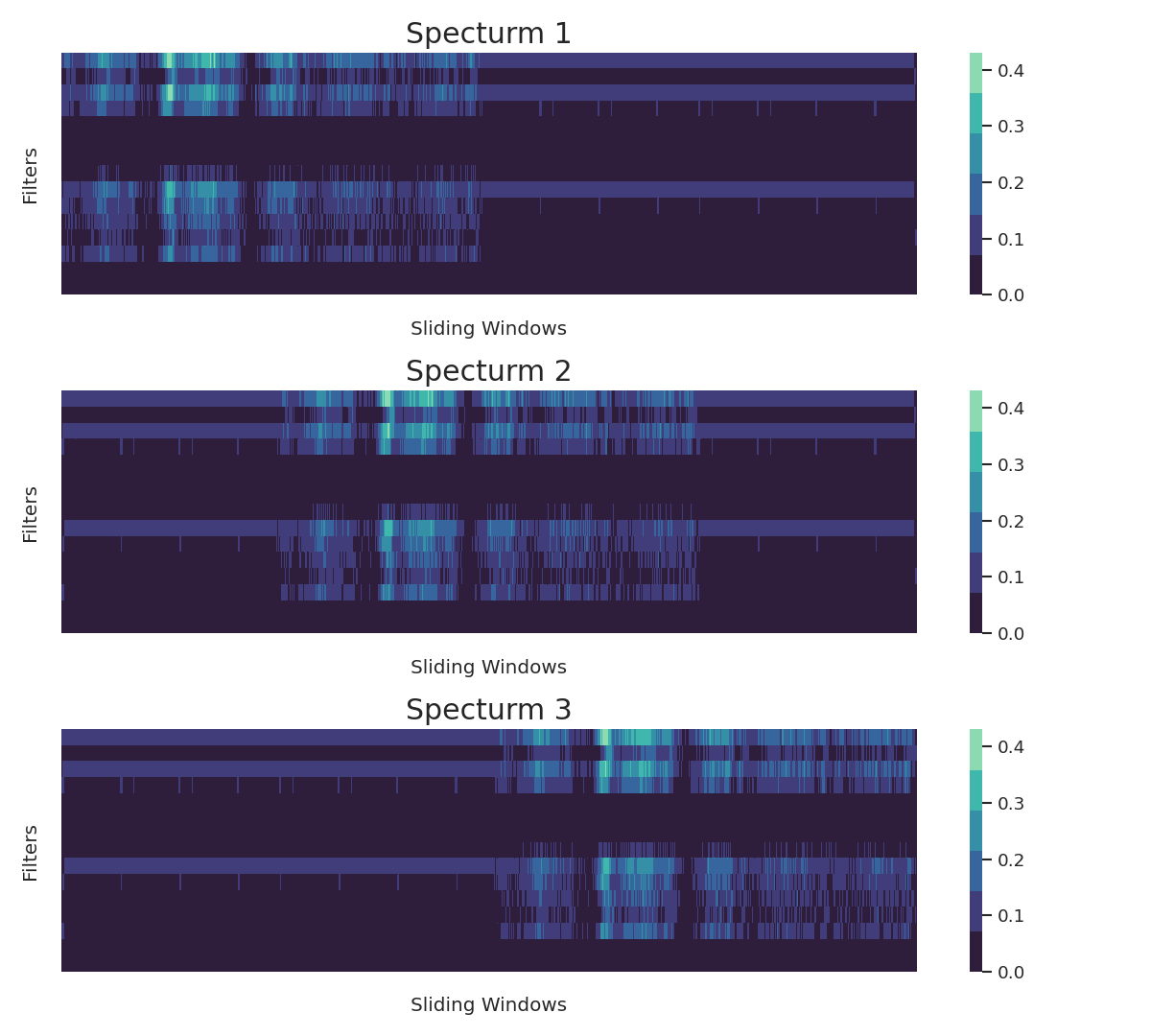}
         \caption{}
     \end{subfigure}
     \begin{subfigure}[b]{0.325\textwidth}
         \centering
         \includegraphics[width=\textwidth]{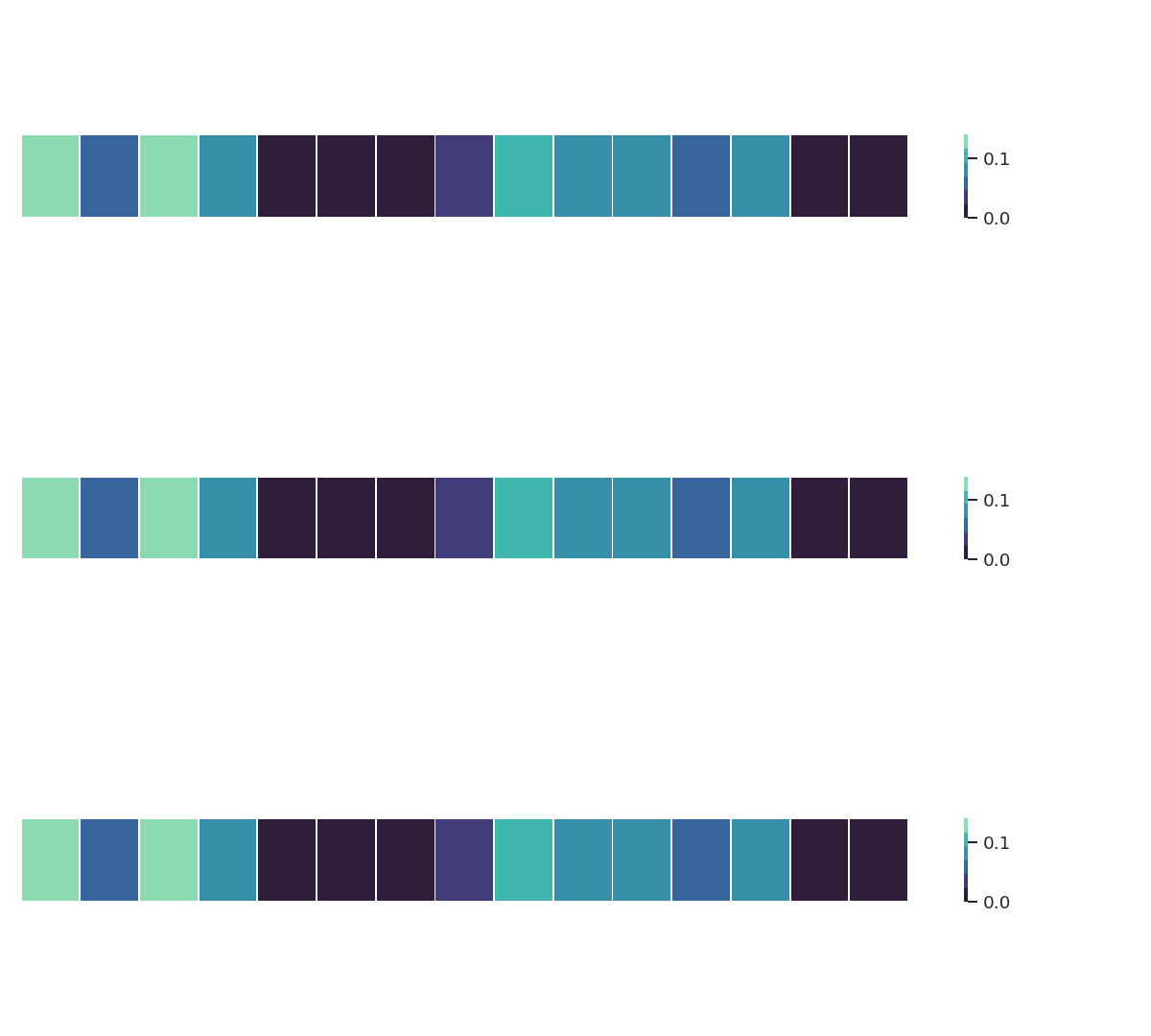}
         \caption{}
     \end{subfigure}
     \caption{Analysis of CNN for a Raman spectrum input.}
     \label{fig:cnn_raman}
\end{figure}

The example clearly shows the suitability of applying CNNs for the time-series signal data. Since the pattern of Raman spectrum closely resembles the pattern of signal data, it is trivial to use CNNs to analyze Raman spectra. The sparse connectivity and parameter sharing truly helps in this regard, as it makes the computation simpler and less prone to overfitting. However, the equivariance to translation property which proved vital for analyzing signals, is not useful when working with the Raman spectra as follows. The Raman spectrum is plotted as intensity vs Raman shifts. Thus the concept of equivariant translation is not applicable in this case, because similar intensity at different Raman shifts implies a completely different meaning. For example, in Fig. \ref{fig:cnn_raman}, we present a similar scenario (as in Fig. \ref{fig:cnn_sig}) but with the Raman spectrum as input. It can be seen that, we have three completely different Raman spectrums, having similar patterns in different Raman shifts. But the CNN model, due to the translational equivariance, treats them as shifted inputs and generates the same output for all of them (Fig. \ref{fig:cnn_raman}c).

The above examples resurface the question of suitability and efficacy of CNN in Raman spectrum analysis. Attempts have been made to address this question in prior works by using  deeper networks (broadening the field of vision thereby) along with discarding pooling layers (with a goal to preserve the locations of spectral peaks) \cite{ho2019rapid}. On the contrary we can use traditional neural networks or machine learning models which are capable of keeping track of the exact locations. However, they succumb to the curse of dimensionality \cite{verleysen2005curse}, due to the long length of the Raman spectrum. Although PCA has been mostly used to reduce that \cite{lussier2020deep}, still it feels less intuitive to model spectrum inputs using hyperplane optimization, which is more suitable for tabular data. As the spectra are apparently correlated in neighboring Raman shifts, this rather motivates us to use sparse connectivity through kernel-like operations in CNN instead, which is also supported by the reduced accuracy in classical models \cite{liu2017deep,ho2019rapid}.

This brings us to the dilemma of whether to use CNN or classical machine learning pipelines. On one hand, CNNs help in analyzing Raman spectra for the benefits of sparse connectivity (considering the correlation in neighboring Raman shifts) and parameter sharing (reducing the number of parameters and overfitting), but loses their applicability due to translational equivariance. On the other hand, traditional models are free from translational equivariance, but they cannot comprehend the correlation in the neighboring Raman shifts and suffers from the curse of dimensionality as trying to optimize too many parameters at once. In the sequel, we overcome this dilemma by proposing a middle ground between the two approaches, fusing the best of the both worlds. We propose to use shifted multi-layer perceptrons (MLPs) to analyze shifted windows of Raman spectra. This facilitates sparse connectivity as the shifted MLP layers mimic kernels and only analyze a part of the input. Moreover, parameter sharing is redundant in this regard, as the kernel-like operations of a particular kernel are only performed in one part and not elsewhere. Finally, this also eliminates the issues with translation equivarience, as for differnt locations or different shifted windows we have different MLP layers. Thus, 

Mathematically, a typical 1D CNN operation can be simplified as,

\begin{equation}
    y(i) = \sigma(\sum _h x(i+h) k(h) + b)
\end{equation}

Here, $x$ is the 1D input, $y$ is the output, $k$ is a learned kernel, $b$ is the bias term and $\sigma$ is a nonlinear operation. The same kernel $k$ is applied everywhere thus the translational equivarience is achieved.

On the contrary, our proposed modification from using a MLP,

\begin{equation}
    y(i) = \sigma(W_{f(i)}^T x  + b) \equiv \sigma(\sum _h x(i+h) k_{f(i)}(h) + b)
\end{equation}

Here, the dot product $W^Tx$ is mathematically equivalent to an 1D convolutional operation with proper relation between the weight matrix $W$ and kernel $k$. In addition since we are using sliding windows, the weight matrix $W_{f(i)}$ and kernel $k_{f(i)}$ depends on the location, i.e., the value of $i$.

\section{Proposed Architecture}

On the basis of the above discussion, we are presenting here a novel network architecture, RamanNet. As mentioned in the previous section, we mimic the convolution operation using multi-layer perceptrons or so-called densely connected neural network layers. The input Raman spectrum is broken into overlapping sliding windows of length $w$ and step size $dw$, and each of them is passed to a different dense block with $n_1$ neurons each. This ensures sparse connectivity and reduces the risk of overfitting. Furthermore, this configuration somewhat resembles a convolution operation without translation, thus we can consider the features extracted at the neurons the same way as we would consider features learned from kernels. 

The features from all the dense blocks are concatenated together and a dropout $dp_1$ is applied. These concatenated features are again summarized using another dense layer with $n_2$ neurons. The outputs from the summarization are regularized with a dropout of $dp_2$.

Finally, we compute $n_f$ features from the regularized outputs using another dense layer with $n_f$ neurons. We name this layer Embedding Layer. Raman spectra are known to be noisy with low signal-to-noise ratio (SNR) \cite{ho2019rapid}, which often leads to less separation between the classes. Therefore, in order to improve this we introduce Triplet Loss \cite{schroff2015facenet} as an auxiliary loss. Triplet Loss is defined using Eucleidian distance, $f$, between an anchor $A$, positive example $P$ and negative example $N$ as,

\begin{equation}
    L(A,P,N) = max(||f(A)-f(P)||^2 - ||f(A)-f(N)||^2 + \alpha, 0)
\end{equation}

Here, $\alpha$ is a margin term to ensure that the model learns non-trivial information.

Using triplet loss we obtain a well-separated embedding space. The embeddings are finally used to predict the classes of input using the Softmax activation function and a dropout of $dp_3$. The network is thus updated using a combination of triplet and cross-entropy loss.

\begin{equation}
    Loss = 0.5 \times \text{triplet loss} + 0.5 \times \text{cross-entropy loss}
\end{equation}

Other than the output layer, all the layers uses LeakyRELU activation function \cite{maas2013rectifier} and are batch normalized \cite{ioffe2015batch}. In our experiments we used the hyper-parameters as follows:

\begin{table}[ht]
\centering
\caption{Hyper-parameters of RamanNet}
\begin{tabular}{|l|c|c|c|c|c|c|c|c|}
\hline
Hyper-parameter & w  & dw & $n_1$ & $n_2$ & $n_f$ & $dp_1$ & $dp_2$ & $dp_3$ \\ \hline
Value      & 50 & 25 & 25    & 512   & 256   & 0.5    & 0.4    & 0.25   \\ \hline
\end{tabular}
\end{table}

\begin{figure}[ht]
    \centering
    \includegraphics[width=\textwidth]{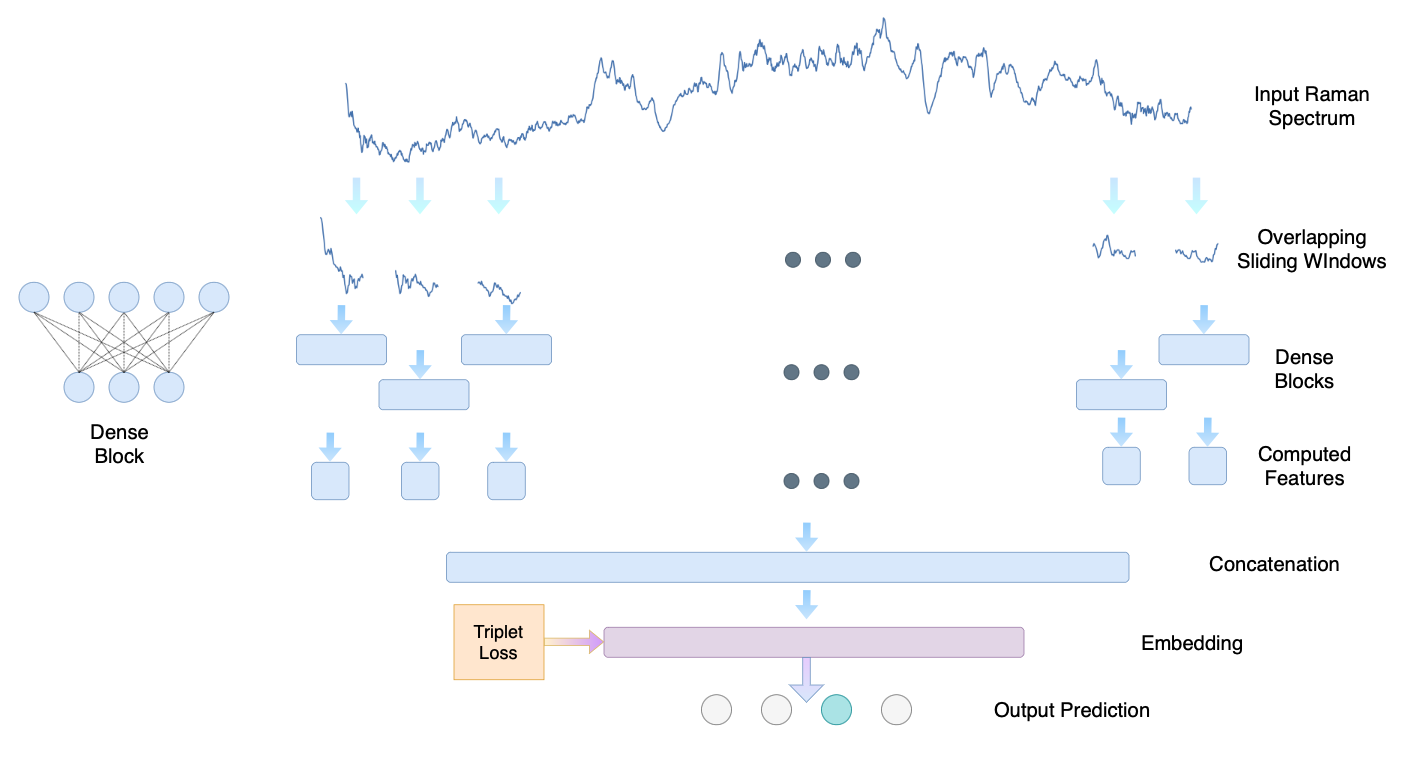}
    \caption{RamanNet Architecture}
    \label{fig:ramannet}
\end{figure}

A simplified diagram of the RamanNet has been presented in Fig. \ref{fig:ramannet}.

\section{Datasets}

One particular limitation when working with deep learning architectures for Raman spectrum analysis is the lack of sufficient public benchmark datasets \cite{lussier2020deep}. Although there have been several of recent works utilizing deep learning models for Raman spectrum analysis, the datasets are mostly proprietary or private \cite{yan2019tongue,thrift2019surface}. From an elaborate review of the recent works on Raman Spectrum analysis, we selected 4 publicly available datasets for our analysis and benchmarking.

\subsection{Covid Dataset}

We used the publicly available data \cite{covid_data} from the recent work \cite{yin2021efficient}, which acts as a pilot study of primary screening of COVID-19 by Raman spectroscopy. This dataset contains a total of 177 serum samples collected from 63 COVID-19 patients, 59 suspected ones and 55 healthy people (i.e.,  control group). The COVID-19 group was recruited at the Chengdu Public Health Clinical Medical Center, and it includes 58 symptomatic and 5 asymptomatic patients. The suspected group demonstrated `flu' like symptoms but were tested negative using RT-PCR tests. For all the subjects, 1-hour repose of blood sampling the serum was extracted by centrifuging at 3000 $rpm$ for 10 minutes, and was stored at 4 \textcelsius. Later, a single-mode laser diode with 785 $nm$ wavelength and 100 $mW$ power was used for Raman excitation. The laser power applied on the sample was measured around 70 $mW$ and the spectra were recorded in the range of 600-1800 $cm^{-1}$.

\subsection{Melanoma Dataset}

Erzina et al. \cite{erzina2020precise} used Surface Enhanced Raman Spectroscopy (SERS) to detect skin melanoma. Healthy residual skin and skin melanoma metastasis were collected. From cell line and primary culture 12 categories of samples were considered, each having 8-9 samples. The SERS spectra were measured using a ProRaman-L spectrometer at 785 $nm$ excitation wavelength and 33 $mW$ power. AuMs, functionalized by ADT-NH2, ADT-COOH or ADT-(COOH)$_2$  was used to measure the SERS spectra, i.e., 3 different spectra were obtained for each sample. Finally, the background was removed using smoothing algorithms and the recorded spectra were normalized to an intensity value of 0 to 1. The range of the recorded spectra is 100-4278 $cm^{-1}$.

\subsection{Mineral Dataset}

Mineral substance identification is another popular application of Raman spectroscopy. Among the various mineral databases, we selected the RRUFF database \cite{lafuente20151}. The RRUFF project aims at curating the most comprehensive set of high-quality spectral data from well-characterized minerals, comprising Raman spectra, X-ray diffraction and information from chemistry. The database contains Raman spectra of various minerals at different configurations, i.e. varied wavelengths and orientations, accompanied by various kinds of processing. Furthermore, the Raman spectra computed from the different materials are hardly consistent, e.g., the ranges of Raman shifts are also different. In order to alleviate such irregularities, we have only considered spectra computed at 432 $nm$, which is the majority. Moreover, we have only collected the raw spectra, i.e., which were not processed any way. Since the recorded spectra have different ranges of Raman shift, we have cropped the region that is common in all the spectra and used cubic spline to interpolate the locations if necessary.The resultant spectra cover the range 280 - 4237 $cm^{-1}$ and then, min-max normalization was performed. Finally, we take the mineral classes with at least 10 samples, which reduces the database to twenty mineral classes and use them to evaluate the proposed model.

\subsection{Bacteria Dataset}

We collected the Bacteria-ID dataset from \cite{ho2019rapid}, where the potential of Raman spectroscopy in label-free bacteria detection was investigated. This dataset consists of 30 bacterial and yeast isolates, including multiple isolates of Gram-negative and Gram-positive bacteria. The dataset is organized into a reference training dataset, reference fine-tuning set and test set. The fine-tuning dataset is used to account for the changes in measurement caused by optical system efficiency degradation. The training dataset contains 2000 spectra for each of the 30 isolates, whereas the fine-tuning and test set contains 100 spectra for each isolate. The isolates were cultured on blood agar plates sealed with Parafilm and stored at 4 $^\circ C$.  The Raman spectra were generated using Horiba LabRAM HR Evolution Raman microscope, with 633 $nm$ illumination at 13.17 $mW$ along with a 300 $l/mm$ grating. The spectra were computed at  1.2 $cm^{-1}$ dispersion to simultaneously maximize signal strength and minimize background signal. The recorded spectra were normalized to the intensity of $0-1$, covering the spectral range between 381.98-1792.4 $cm^{-1}$.

\section{Experimental Setup}

The experiments have been conducted in a server computer with Intel Xeon @2.2GHz CPU, 24 GB RAM, and NVIDIA TESLA P100 (16 GB) GPU. We implemented the RamanNet architecture using Tensorflow \cite{abadi2016tensorflow}. The codes are available in the following Github repository.

\begin{center}
    \url{https://github.com/nibtehaz/RamanNet}    
\end{center}

In the following subsections, we briefly explain the experimental protocols and evaluation procedures adopted for the different tasks. We have tried to mimic the corresponding baseline papers to ensure a fair comparison. As mentioned earlier, we were only compelled to follow a different evaluation scheme for the Mineral and Melanoma dataset, thus we reproduced that baseline model's output following our exact protocol.

\subsection{Covid Dataset}

We followed the same evaluation method as presented in \cite{yin2021efficient}. In this work, the authors conducted a ``blind'' validation. They randomly divided the whole dataset into training (70\%) and hold-out test set (30\%). In order to further assess the independence of the data over model performance, this process was repeated 50 times and the average values of the metrics were recorded. We followed the same protocol to evaluate RamanNet. In order to avoid overfitting, we used 10\% data from the training set as validation data, but the hold-out test data was left completely independent from the training process and was only used for evaluation. The models were trained for 1000 epochs.

\subsection{Melanoma Dataset}

In the original work, Erzina et al. \cite{erzina2020precise}, performed a 75\%:25\% train-validation split. Therefore, we attempted to follow a similar splitting criterion. Since, the splitting information was not provided, we opted to perform a 4 fold cross-validation instead, as it would also split the data in the same ratio. The authors demonstrated 100\% accuracy using 3 different spectra (AuMs functionalized by ADT-NH2, ADT-COOH or ADT-(COOH)$_2$ respectively) together. However, to make the task difficult, we experimented with using only one type of spectra as input. To ensure a level-playing field, we reproduced the model as described in \cite{erzina2020precise} and evaluated the model with one particular spectrum as input at a time.

\subsection{Mineral Dataset}

Jichao et al. \cite{liu2017deep} used the RRUFF database for mineral classification task. However, they employed the leave one out cross-validation scheme, which is computationally too expensive. In order to reduce the computational requirements, we thus opted for a 5 fold cross-validation instead. In order to compare RamanNet with the model presented by Jichao et al. \cite{liu2017deep}, we implemented their proposed model and used it in our analysis.

\subsection{Bacteria Dataset}

In order to assess RamanNet on the Bacteria-ID dataset, we followed the same training and evaluation procedure as presented in \cite{ho2019rapid}. Similar to their approach, we first pre-trained the model using the reference training dataset, through a 5 fold cross-validation scheme. The five models obtained in this process were then fine-tuned on the fine-tuning dateset, which was spilt into 90\% training and 10\% validation set. The model with the highest accuracy on this validation set was considered and was evaluated on the independent test dataset. The models were trained for 100 and 250 epochs respectively on the reference training and fine-tuning set.

\section{Results}

\subsection{RamanNet Consistently Outperforms Existing Models}

\subsubsection{COVID-19 Dataset}

In \cite{yin2021efficient}, an SVM model was developed to distinguish the different categories, namely healthy, suspected and COVID-19 patients. Instead of working with the entire spectrum, wave points with significant differences in the ANOVA test were selected. Thus, a statistically sound feature reduction was performed and the reduced featureset was used as the input to the SVM model. On the contrary, RamanNet takes the entire spectrum as input and adaptively finds the significant region therein. 

The average performance over 50 random trials for the different tasks is presented in Table \ref{tbl:covid_res}, here we present the accuracy, sensitivity and specificity values. Among the 3 tasks, differentiating between suspected and healthy subjects seems to be the most challenging one, as is evident from the inferior performance of SVM (all metrics $\leq$ 70\%). RamanNet, on the other hand, performed comparatively better in this task. Notably, RamanNet improved accuracy and specificity by 13\% and 21\% respectively. RamanNet also achieved an improved sensitivity score.

On the other two tasks the SVM model performed comparatively  (than its own performance in the first task). However, our proposed RamanNet consistently outperformed SVM in all these two  tasks as well. Most promisingly, RamanNet improved sensitivity greatly in both of the tasks. For this problem sensitivity is crucial as we need to correctly detect the Covid-19 patients. This improvement in sensitivity did not come at any cost of specificity, rather the specificity has also been improved compared to the SVM model.

\begin{table}[ht]
\centering
\caption{Results on COVID-19 Dataset}
\label{tbl:covid_res}
\begin{tabular}{|c|c|c|c|}
\hline
\multicolumn{4}{|c|}{COVID-19 vs Suspected} \\ \hline
Method & Accuracy & Sensitivity & Specificity \\ \hline
SVM & $87 \pm 5$ & $89 \pm 8$ & $86 \pm 9$ \\ \hline
RamanNet & $93 \pm 3$ & $97 \pm 4$ & $90 \pm 6$ \\ \hline
\multicolumn{4}{|c|}{COVID-19 vs Healthy} \\ \hline
Method & Accuracy & Sensitivity & Specificity \\ \hline
SVM & $91 \pm 4$ & $89 \pm 7$ & $93 \pm 6$ \\ \hline
RamanNet & $95$ & $95 \pm 4$ & $96 \pm 3$ \\ \hline
\multicolumn{4}{|c|}{Suspected vs Healthy} \\ \hline
Method & Accuracy & Sensitivity & Specificity \\ \hline
SVM & $69 \pm 5$ & $70 \pm 9$ & $66 \pm 9$ \\ \hline
RamanNet & $82 \pm 6$ & $77 \pm 15$ & $87 \pm 11$ \\ \hline
\end{tabular}
\end{table}

\subsubsection{Melanoma Dataset}

As described previously, we perform a 4 fold cross-validation on the melanoma dataset. Following the original evaluation as performed by Erzina et al. \cite{erzina2020precise}, we consider all the three different types of spectra simultaneously as input. In addition, we perform a difficult version of the problem by taking only one type of spectra as input at a time. The results are presented in Table \ref{tbl:res_melanoma}.

\begin{table}[ht]
\caption{Results on Melanoma Dataset}
\label{tbl:res_melanoma}
\begin{tabular}{|c|c|c|c|c|c|c|c|c|c|c|}
\hline
 & \multicolumn{2}{c|}{-NH$_2$} & \multicolumn{2}{c|}{-(COOH)$_2$} & \multicolumn{2}{c|}{-COOH} & \multicolumn{2}{c|}{All} & \multicolumn{2}{c|}{\#Parameters} \\ \hline
Fold & RamanNet & CNN & RamanNet & CNN & RamanNet & CNN & RamanNet & CNN & RamanNet & CNN \\ \hline
1 & \textbf{100} & 97.42 & \textbf{100} & 100 & \textbf{99.35} & 94.19 & 100 & 100 & \multirow{4}{*}{1.3 M} & \multirow{4}{*}{25.7M} \\ \cline{1-9}
2 & \textbf{99.35} & 96.13 & \textbf{99.35} & 98.71 & \textbf{98.71} & 96.12 & 100 & 98.71 &  &  \\ \cline{1-9}
3 & \textbf{100} & 95.45 & \textbf{100} & 98.05 & \textbf{99.35} & 87.01 & 100 & 100 &  &  \\ \cline{1-9}
4 & \textbf{100} & 97.40 & \textbf{100} & 98.70 & \textbf{96.75} & 96.10 & 100 & 99.35 &  &  \\ \hline
\end{tabular}
\end{table}

From the results, it is evident that although the CNN model manages to achieve perfect 100\% accuracy (in 2 folds out of 4) when given 3 spectra as input, the accuracy falls when a single spectrum is given as input. This drop in performance can be explained by the loss of information, when working with a single spectrum. For different functionalizations of AuMs, the sample is observed from a different point of view and different information is obtained. Thus, when working with a reduced number of spectra, insightful information is likely to get lost and that negatively affects the performance. Although for the -(COOH)$_2$ as input the accuracy of the CNN model stays above 98\%, it falls below 97\% for the other two input spectra CNN model.

RamanNet on the other hand seems to consistently outperform the CNN model for both when all the 3 spectra are considered together or separately. RamanNet not only consistently achieved 100\% accuracy with all the 3 spectra as input, but also with -NH$_2$ and -(COOH)$_2$ individual spectra as input separately. Only when -COOH spectra is used as input, the performance was not up to the mark, but still the performance was superior to the CNN. All these improvements become more significant when we compare the number of the parameters of the two models. The CNN model consists of 25.7 M parameters whereas RamanNet has only 1.3 M parameters ($\sim 5\%$). Thus, RamanNet is not only more accurate, but it is also computationally efficient at the same time.

\subsubsection{Mineral Dataset}

As described in the previous sections, the mineral dataset consists of 20 mineral classes and in order to compare with the model proposed by \cite{liu2017deep}, we implement their model and perform a five-fold cross-validation. The results are presented in Table \ref{tbl:res_mineral}.

\begin{table}[ht]
\caption{Results of Mineral Dataset}
\label{tbl:res_mineral}
\begin{tabular}{|c|c|c|c|c|c|c|c|c|}
\hline
 & \multicolumn{2}{c|}{Top 1 Accuracy} & \multicolumn{2}{c|}{Top 5 Accuracy} & \multicolumn{2}{c|}{Top 10 Accuracy} & \multicolumn{2}{c|}{\#Parameters} \\ \hline
 & RamanNet & CNN & RamanNet & CNN & RamanNet & CNN & RamanNet & CNN \\ \hline
Fold 1 & \textbf{87.5} & 81.25 & \textbf{95.83} & 87.5 & \textbf{100} & 93.75 & \multirow{5}{*}{1.3M} & \multirow{5}{*}{6.6M} \\ \cline{1-7}
Fold 2 & \textbf{93.75} & 85.42 & 97.92 & 97.92 & \textbf{100} & 97.92 &  &  \\ \cline{1-7}
Fold 3 & \textbf{89.58} & 79.17 & \textbf{93.75} & 89.58 & \textbf{93.75} & 91.67 &  &  \\ \cline{1-7}
Fold 4 & \textbf{97.92} & 85.42 & \textbf{100} & 91.67 & \textbf{100} & 93.75 &  &  \\ \cline{1-7}
Fold 5 & \textbf{85.12} & 63.83 & \textbf{93.62} & 78.72 & \textbf{95.74} & 85.10 &  &  \\ \hline
\end{tabular}
\end{table}

Here, we have presented the Top 1, Top 5 and Top 10 accuracy metrics, which are popularly used for multiclass classification problems. In Top X accuracy score, we check whether the top X predictions of the model matches with the ground truth. It is evident that RamanNet consistently outperforms the previous state-of-the-art CNN model. For Top 1 Accuracy, the improvement is more prominent, nevertheless RamanNet performs better in regards to the other metrics as well. Another point worth mentioning is that, even in this case the CNN model is heavier comparatively (6.6 M parameters vs. 1.3 M parameters for RamanNet).

\subsubsection{Bacteria Dataset}

\begin{figure}[ht]
     \centering
     \begin{subfigure}[b]{0.325\textwidth}
         \centering
         \includegraphics[width=\textwidth]{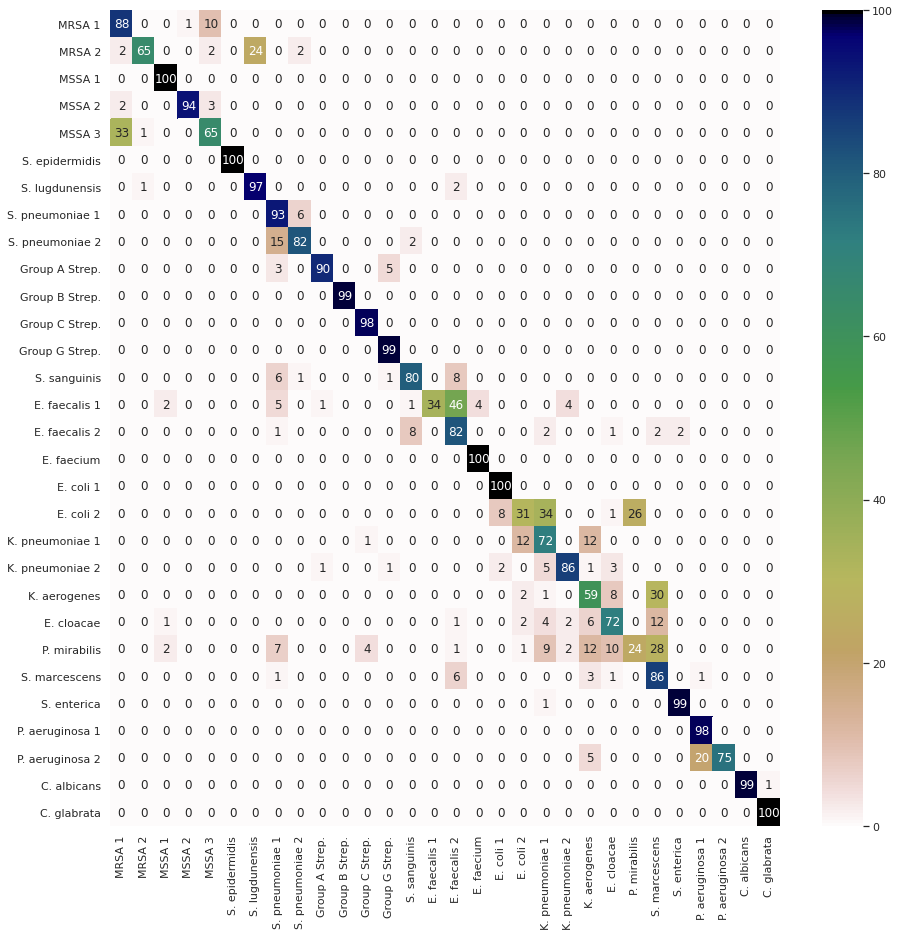}
         \caption{25 Layer ResNet}
         \label{fig:bacteria_resnet}
     \end{subfigure}
     \begin{subfigure}[b]{0.325\textwidth}
         \centering
         \includegraphics[width=\textwidth]{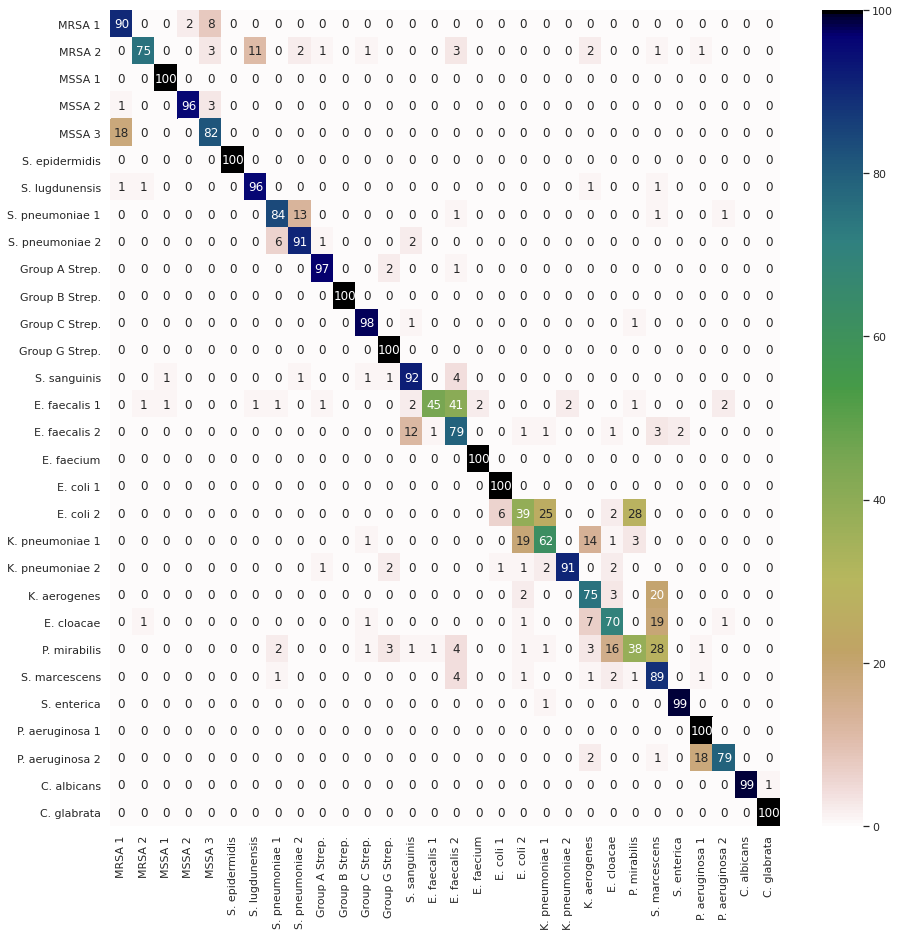}
         \caption{RamanNet}
         \label{fig:bacteria_ramannet}
     \end{subfigure}
     \begin{subfigure}[b]{0.325\textwidth}
         \centering
         \includegraphics[width=\textwidth]{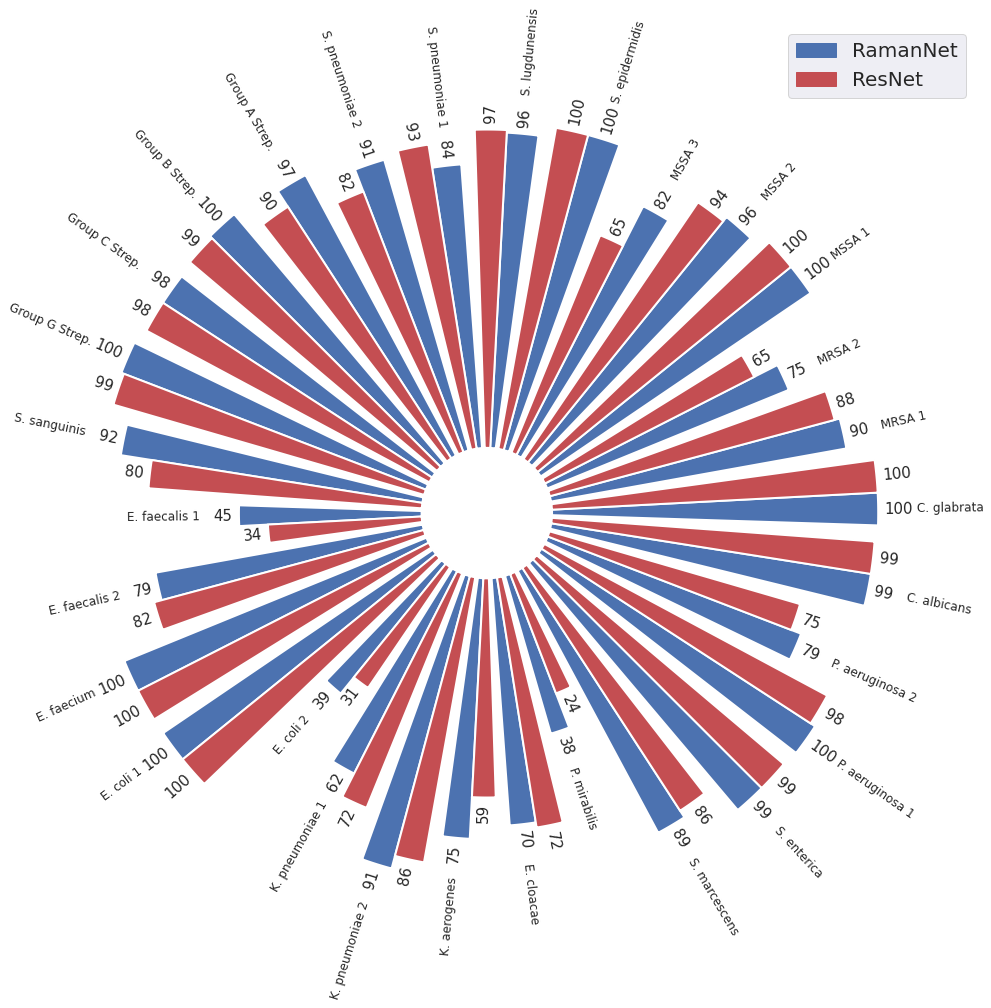}
         \caption{Comparison}
         \label{fig:bacteria_comparison}
     \end{subfigure}
     \caption{Confusion matrix for Bacteria Dataset using 25-layer ResNet (a), RamanNet (b) and their comparison (c).}
\end{figure}

In the original work \cite{ho2019rapid}, the authors used a 25 layer deep residual convolutional neural network to classify the bacteria isolate Raman spectrum to one of the 30 classes. The proposed model achieved an average accuracy of 82.2\%, and the resulting confusion matrix is presented in Fig. \ref{fig:bacteria_resnet}. The authors also experimented with common and popular classifiers like Logistic Regression (LR) and Support Vector Machine (SVM) as baselines, but those models could only reach 75.7\% and 74.9\% accuracy respectively.

RamanNet on the other hand manages to achieve an average accuracy of 85.5\% on this dataset, despite only having 3 hidden layers. As shown in the confusion matrix (Fig. \ref{fig:bacteria_ramannet}), the number of misclassifications has reduced. Although erroneous predictions still exist, most Gram-positive and Gram-negative bacteria have been misclassified as Gram-positive and Gram-negative bacteria respectively, and the errors are also mostly confined within the same genus, as analyzed in \cite{ho2019rapid}. Although this 3.3\% improvement may seem minor, this is achieved using a much shallower network (3 layers vs 25 layers). Furthermore, the improvements become more apparent when we compare the individual classifications next to each other. As presented in Fig. \ref{fig:bacteria_comparison}, RamanNet achieves either equal or better accuracy for 25 out of 30 bacteria isolate classes. For the other two isolate classes, ResNet is only better with a small margin. On the contrary, in the cases where RamanNet performed better it surpassed ResNet with a higher margin (e.g., Methicillin sensitive \textit{Staphylococcus aureus} (MSSA 3), \textit{Proteus mirabilis}, \textit{Klebsiella aerogenes} etc.).

\subsection{Triplet Loss Improves Dimensionality Reduction}

The majority of Raman spectra analysis works have been based on classical machine learning methods. Therefore, it has always been crucial to reduce the feature-space. In this regard principal component analysis (PCA) has been the most prominent method for feature reduction, to an extent that 34 out of recent 52 papers used PCA \cite{lussier2020deep}. Therefore, in the Raman spectra analysis community, feature selection and/or reduction is almost equally important as accurate classification.

Therefore, in order to perform the task of dimensionality reduction of Raman spectra, we have put focus on embedding generation capability of RamanNet. In addition to calibrating the embeddings learned by RamanNet from the class labels through backpropagated cross-entropy loss, we also include triplet loss in the embedding layer. This allows to simultaneously minimizing intraclass distance while maximizing interclass distance.

\begin{figure}[ht]
     \centering
     \begin{subfigure}[b]{0.99\textwidth}
         \centering
         \includegraphics[width=\textwidth]{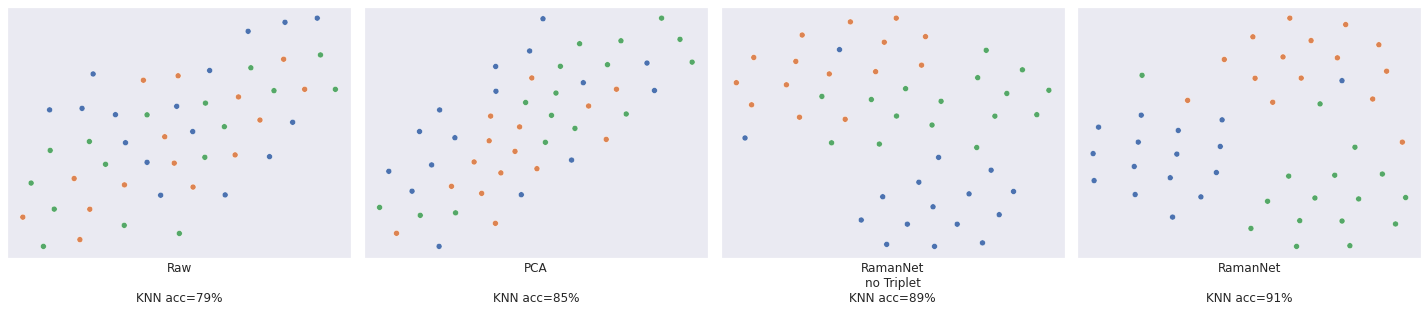}
         \caption{Covid Dataset}
     \end{subfigure}
     \begin{subfigure}[b]{0.99\textwidth}
         \centering
         \includegraphics[width=\textwidth]{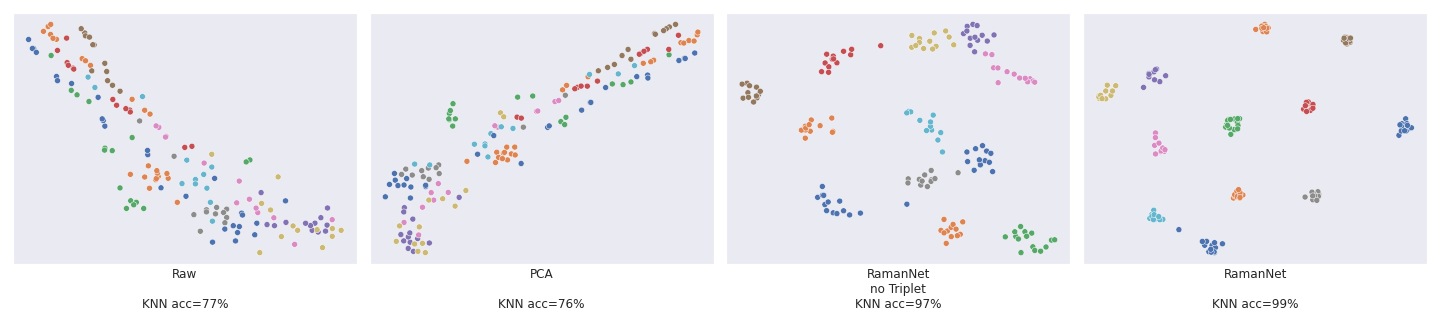}
         \caption{Melanoma Dataset}
     \end{subfigure}
     \begin{subfigure}[b]{0.99\textwidth}
         \centering
         \includegraphics[width=\textwidth]{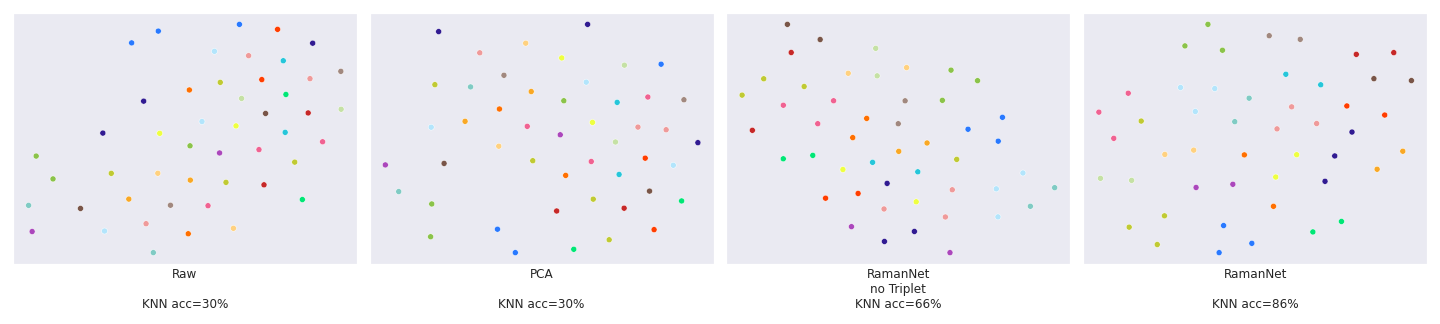}
         \caption{Mineral Dataset}
     \end{subfigure}
     \begin{subfigure}[b]{0.99\textwidth}
         \centering
         \includegraphics[width=\textwidth]{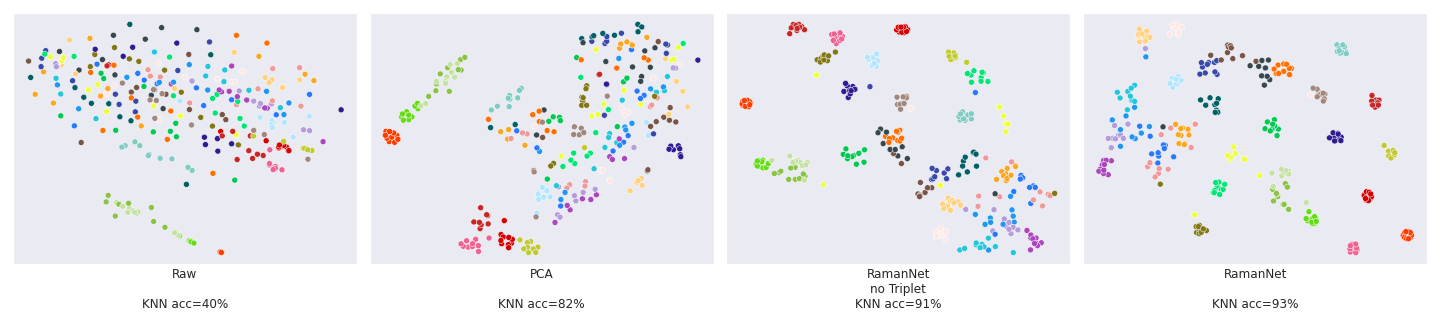}
         \caption{Bacteria Dataset}
     \end{subfigure}
     \caption{t-SNE embeddings of the raman spectra of different datasets for different feature representations. Here, 2-dimensional t-SNE embeddings have been computed from the original raw spectrum, PCA features, RamanNet embeddings without and with triplet loss respectively. In addition, we also report the accuracy of a simple KNN classifier for the individual feature representations.}
     \label{fig:tsne}
\end{figure}

In order to assess the quality of the embeddings generated by RamanNet, we compare the RamanNet embeddings with PCA and the original raw spectrum. We also train a version of RamanNet without the triplet loss, to analyze the contribution of triplet loss. For qualitative analysis of the class separation obtained from such feature reduction, we plot 2-dimensional T-distributed Stochastic Neighbor Embedding (t-SNE) plots \cite{van2008visualizing}. From fig. \ref{fig:tsne}, it can be observed that RamanNet embeddings are significantly superior than PCA or the original spectrum. Furthermore, RamanNet trained with triplet loss produces better embedding than training the model without this loss. 

\clearpage

Since, t-SNE is an approximate low-dimensional representation, we train a simple KNN models with 15 neighbors and perform classification, as a mean of quantitative evaluation. Even in this case, it is evident that RamanNet trained with triplet loss is capable of the most desired feature representation. The results are presented in Table \ref{tbl:knn}.

\begin{table}[ht]
\caption{Simple KNN model accuracy for various feature representations}
\label{tbl:knn}
\centering
\begin{tabular}{|cl|cccc|}
\hline
\multicolumn{2}{|l|}{\multirow{2}{*}{}} & \multicolumn{4}{c|}{Dataset}                                                                         \\ \cline{3-6} 
\multicolumn{2}{|l|}{}                  & \multicolumn{1}{c|}{Covid} & \multicolumn{1}{c|}{Melanoma} & \multicolumn{1}{c|}{Mineral} & Bacteria \\ \hline
\multicolumn{1}{|c|}{\multirow{4}{*}{\begin{tabular}[c]{@{}c@{}}Feature\\ Representation\end{tabular}}} &
  Raw &
  \multicolumn{1}{c|}{79\%} &
  \multicolumn{1}{c|}{77\%} &
  \multicolumn{1}{c|}{30\%} &
  40\% \\ \cline{2-6} 
\multicolumn{1}{|c|}{}    & PCA         & \multicolumn{1}{c|}{85\%}  & \multicolumn{1}{c|}{76\%}     & \multicolumn{1}{c|}{30\%}    & 82\%     \\ \cline{2-6} 
\multicolumn{1}{|c|}{} &
  \begin{tabular}[c]{@{}l@{}}Without\\ triplet loss\end{tabular} &
  \multicolumn{1}{c|}{89\%} &
  \multicolumn{1}{c|}{97\%} &
  \multicolumn{1}{c|}{66\%} &
  91\% \\ \cline{2-6} 
\multicolumn{1}{|c|}{}    & RamanNet    & \multicolumn{1}{c|}{91\%}  & \multicolumn{1}{c|}{99\%}     & \multicolumn{1}{c|}{86\%}    & 93\%     \\ \hline
\end{tabular}
\end{table}

Fisher discriminant ratio (FDR) \cite{RAJOUB202029} is another measure of class separability. FDR provides a score of the features based on the centroids and spreads of the classes. For a dataset with $C$ classes, each class $i$ having $n_i$ samples, suppose the mean and standard deviation values of a feature $x_r$ is $\mu_i$ and $\sigma_i$ respectively for class $i$. If the global mean and standard deviation of feature $x_r$ is $\mu$ and $\sigma$ respectively, then the FDR value for that feature is defined as:

\begin{equation}
    FDR_r = \frac{\sum_{i=1}^C n_i (\mu_i-\mu)^2}{\sum_{i=1}^C n_i \sigma_i^2}
\end{equation}

\begin{figure}[ht]
     \centering
     \begin{subfigure}[b]{0.24\textwidth}
         \centering
         \includegraphics[width=\textwidth]{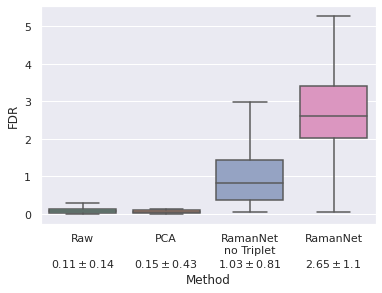}
         \caption{Covid Dataset}
     \end{subfigure}
     \begin{subfigure}[b]{0.24\textwidth}
         \centering
         \includegraphics[width=\textwidth]{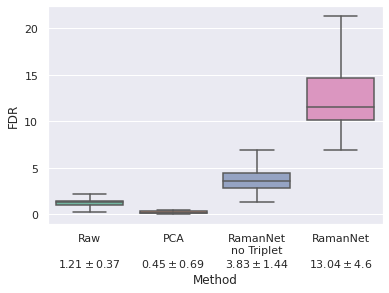}
         \caption{Melanoma Dataset}
     \end{subfigure}
     \begin{subfigure}[b]{0.24\textwidth}
         \centering
         \includegraphics[width=\textwidth]{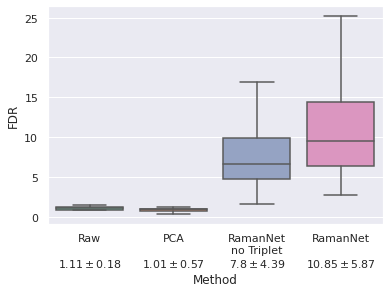}
         \caption{Mineral Dataset}
     \end{subfigure}
     \begin{subfigure}[b]{0.24\textwidth}
         \centering
         \includegraphics[width=\textwidth]{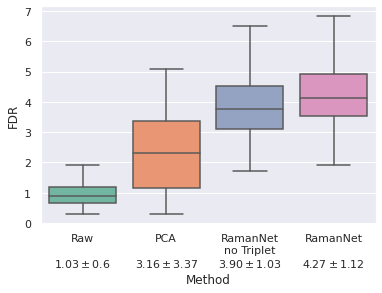}
         \caption{Bacteria Dataset}
     \end{subfigure}
     \caption{Fisher discriminant ratio (FDR) scores for different feature representations.}
     \label{fig:fdr}
\end{figure}

Higher value of FDR indicates better class separability, whereas a lower value means that there exists overlaps between the classes in respect to that particular feature. In order to assess the learned features quantitatively, therefore we compute the FDR values of the learned features and compare the scores with PCA and raw spectra. The comparisons are presented in Fig. \ref{fig:fdr}. It is evident that RamanNet trained with triplet loss generates features with high FDR scores consistently. In Mineral and Bacteria dataset the improvement may appear less, this is because when the number of classes increase the notions of inter and intraclass distances gets a bit relaxed.

\subsection{Model Interpretation}

Interpretability has been one of the focuses of deep learning research in recent years \cite{zhang2018visual}. Deep learning models are competent function approximators and given a sufficient amount of data, they are capable of modelling almost any complex functions. With this potential, also comes the concern of what the model is actually learning from the data. The model can learn actual significant information and perform prediction accordingly, or it can merely learn from the noises and get confused from various confounding factors instead. Therefore, it is imperative to investigate the model interpretability and examine what the model is learning from.

For convolutional neural networks, we can use various methods like saliency maps \cite{ibtehaz2021edith} or score-CAM \cite{rahman2020reliable} methods for model interpretation. However, for multi-layer perceptrons, it is non-trivial to do so. The various visualization methods have been designed based on CNNs and they cannot be directly translated to MLPs.

\begin{figure}[ht]
    \centering
    \includegraphics[width=\textwidth]{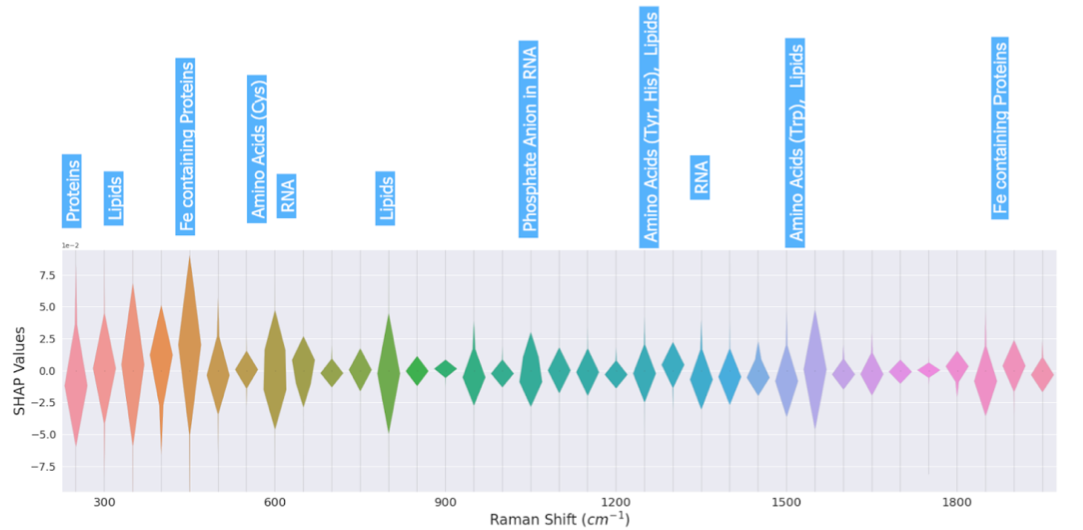}
    \caption{SHAP values for RamanNet}
    \label{fig:shap}
\end{figure}

SHAP (SHapley Additive exPlanations) \cite{NIPS2017_7062} is a game-theoretic approach to explain the output of machine learning models. The biggest advantage of using SHAP is that it is model agnostic, thus it can be used to analyze any machine learning model. We extracted the features extracted from different moving windows of RamanNet and trained the identical top layer of RamanNet using Scikit-Learn MLP implementation, for compatibility reasons of the SHAP implementation \cite{shap_git}. We then computed the SHAP scores of all the -NH$_2$ samples in the melanoma dataset. We choose the melanoma dataset for this experiment because model interpretability is particularly crucial for disease diagnosis and information on significant biological properties related to melanoma was available. The results are presented as a violin plot in Fig. \ref{fig:shap}.

From the violin plot it is evident that certain regions of the spectrum contributes most to the prediction. Moreover, those regions correspond to the actual significant properties of the sample. For example, the region in 400-500 cm$^{-1}$, corresponding to Fe containing proteins, contributes the most to the prediction. After this region the lipids (300-400 cm$^{-1}$) contributes the most, and so on. The unlabeled regions in the figure correspond to less attributed regions, and it is apparent that the model was also aware enough to put less focus on them. All these findings are consistent with the one presented in \cite{erzina2020precise}. Therefore, we can expect that RamanNet is learning significant information from the data and ignoring the noises instead of falling into a conundrum with confounding factors.

\section{Conclusion}

Raman spectroscopy has slowly started to gain more attention with the advances in SERS technology. The gradual decrease of cost and complexity in computing the Raman spectrum, is paving the way to large-scale Raman spectrum data collection for diverse tasks. Therefore, we need suitable machine learning methods to analyze these large-scale Raman spectrum data. However, there has not yet been any model developed for the sole purpose of Raman spectroscopy analysis, motivated and designed based on the unique properties of the Raman spectrum. Recently, existing methods like CNN or SVM have shown success in Raman spectrum analysis, but in this work we presented reasoning that such methods may not be $100\%$ suitable for Raman spectra analysis.

In this work, we present RamanNet, a generalized neural network architecture for Raman spectrum analysis. We take intuitions from the nature of data and design our model accordingly. We propose modifications to the convolutional network behaviors and emulate such operations using multi-layer perceptrons. This adjustment brings the best out of both worlds and it is reflected in our carefully designed experimental evaluation of the model on 4 public datasets. Not only that, the RamanNet outperforms all the state-of-the-art approaches in Raman spectroscopy analysis, it achieves it by adopting much less complexity. Moreover, RamanNet generates embeddings from spectrum which is much better than what is obtained from PCA, the defacto standard in Raman spectrum analysis. Furthermore, we examined the interpretability of RamanNet particularly on a disease dataset and it was revealed that the model is capable of focusing on the (biologically) meaningful information.

The future direction of this research can be manifold. Firstly, we wish to evaluate RamanNet on more large-scale datasets, as they become public. Secondly, we have not performed any preprocessing or background removal of the spectra, we wish to investigate this further with a more dedicated study to infer the denoising capabilities of RamanNet. Last but not the least, we also wish to experiment with multiple particle data to assess if RamanNet is capable of identifying, segmenting and extracting the signatures of the different particles.

\section*{Conflict of Interest}

The authors declare that they have no conflict of
interest.

\section*{Data Availability}

All the datasets used in the experiments are publicly available.

\begin{itemize}
    \item Covid Dataset : \url{https://springernature.figshare.com/articles/dataset/Data_and_code_on_serum_Raman_spectroscopy_as_an_efficient_primary_screening_of_coronavirus_disease_in_2019_COVID-19_/12159924} 
    \item Melanoma Dataset : \url{https://www.kaggle.com/datasets/andriitrelin/cells-raman-spectra}
    \item Mineral Dataset : \url{https://rruff.info} 
    \item Bacteria Dataset : \url{https://github.com/csho33/bacteria-ID.}
\end{itemize}

\section*{Code Availability}
The codes are available in the following Github repository.

\begin{center}
    \url{https://github.com/nibtehaz/RamanNet}    
\end{center}

\section*{Acknowledgements}

This research is financially supported by Qatar National Research Foundation (QNRF), Grant number NPRP12S-0224-190144. The statements made herein are solely the responsibility of the authors.

\bibliographystyle{unsrt}
\bibliography{main.bib}

\end{document}